# Color Image Segmentation Metrics

**Majid Harouni*and Hadi Yazdani Baghmaleki**
*Department of Computer Science, Dolatabad Branch, Islamic Azad University, Isfahan, Iran*
* Email Address: majid.harouni@gmail.com

**Abstract**
An automatic image segmentation procedure is an inevitable part of many image analyses and computer vision which deeply affect the rest of the system; therefore, a set of interactive segmentation evaluation methods can substantially simplify the system development process. This entry presents the state of the art of quantitative evaluation metrics for color image segmentation methods by performing an analytical and comparative review of the measures. The decision-making process in selecting a suitable evaluation metric is still very serious because each metric tends to favor a different segmentation method for each benchmark dataset. Furthermore, a conceptual comparison of these metrics is provided at a high level of abstraction and is discussed for understanding the quantitative changes in different image segmentation results.

## INTRODUCTION

One of the most important current issues in image analysis and computer vision is evaluation of the segmentation methods of the systems. In fact, the task of image segmentation is to produce a separated image into homogenous and distinct regions, which has some information explicit in analyzing the image processing, and contributes to computerized data extraction. Some well-known examples of image segmentation use for image analysis and vision are scene analysis,[1,2] image retrieval and recognition,[3–6] natural or medical image understanding,[7,8] saliency detection,[9] object tracking systems,[10,11] and many more fields of study. The image segmentation methods can be categorized into seven major classes: region-based,[12,13] boundary-based,[14–16] edge detection-based,[17,18] template matching,[19,20] cluster-based,[21–23] threshold-based,[24–27] graph-based,[28,29] and the combination of the mentioned classes.[30–33] In all the working steps of the aforementioned systems, the image segmentation step has a substantial impact on the overall algorithmic performance; therefore, the process of selecting segmentation evaluation metrics should be considered carefully. The key problems with this selection are that the images are captured with poor contrast, low resolution, illumination variability, scale changes, color balances, complex subsurface structure and/or different texture details, various outdoor conditions, and variation of camera calibration and sensitivity. Generally, the human visual system is well formed to segment a complex and nonuniform image background into the desired objects and/or regions. Nevertheless, it is very difficult to segment such images into perceptually distinct regions. Hence, the evaluation of image segmentation results is necessary to effectively carry out the objectives of the research; this can help researchers to identify the strengths and weaknesses of the image segmentation methods, which are of paramount importance. Nevertheless, the measurement of segmentation performance methods and a standard comparison with other methods have always been the main principles for researchers.

Image segmentation evaluation problem has been reviewed substantially, which is still a continuing challenge in determining the best evaluation method with respect to optimality and computation time for the researchers. Besides, several different decision-making processes have been proposed to tackle this challenge. These processes has the following issues: (1) to efficiently exploit the ability of a proposed segmentation method in comparison with other image segmentation methods; (2) to define the proposed method is paramount in different image categories, for example, natural or medical images; and (3) to validate the selected parameterization biases of the proposed method for various image conditions are more effective in this research plan. This entry provides an extensive description of each image segmentation metrics, where a much more methodical study considers that quantitative evaluation methods would usefully complement and extend the qualitative image analysis. In general, image segmentation evaluation methods can be divided into two main categories:[34–37] empirical evaluation and analytical evaluation methods. The empirical methods analyze a segmentation method with respect to its findings and outputs, some of these methods are presented in the works of Bouthemy and François,[1] Said,[38] and Zhang;[39] instead, the analytical methods assess the method based on its complexity, functionality, utilities, etc.[7,21,23,29]

The rest of the entry is organized as follows: Section "Brief Background" presents the related work. Section "Types of Quantitative Metrics" gives an overview of the structure of image segmentation methods and presents the extended hierarchy of different image segmentation evaluation methods. Section "Discussion and Suggestion" provides the comparative performance analysis of various metrics. Finally, Section "Conclusions" concludes this entry.

## BRIEF BACKGROUND

Image segmentation methods have been utilized in computer vision and image understanding to partition an image into its distinct regions.[10,25,26,38–40] A great variety of color image segmentation methods is presented; some are used for general images and some are planned for a specific purpose. Furthermore, the outputs of these methods can perceptually be classified as correct-segmentation, over-segmentation, and under-segmentation.[29,41–44] A color image dataset is rich in both homogeneous color regions and texture; in this case, the texture-based segmentation method would be proposed and the most challenging problems are as follows: (1) to segment the elements of texture regions in various shapes and sizes without losing desired details; (2) to reduce the processing time of the method; and (3) to evaluate the computational efficiency of the different algorithms. In the following section, a brief literature survey of related works on the third challenging problem is summarized. A color–texture image segmentation method was proposed based on a three-dimensional deformable surface model and energy function;[45] a set of quantitative evaluation metrics is applied to measure the quality of the proposed method in natural image scenes, that is, Probabilistic Rand Index (PRI), Normalized Probabilistic Rand (NPR) index, global consistency error (GCE), variation of information (VOI), and boundary displacement error (BDE). In the work of Mridula,[46] a hybrid color texture segmentation model was developed for pixel labeling of images by combining -level co-occurrence matrix feature extraction and Markov random field algorithm; the hybrid model was evaluated using misclassification error (MCE) metrics. In the proposed image segmentation method in Ref. [21] the different pixels or regions were extracted from labeling the pixel-level image features using support vector machine classification and the evaluation of the proposed method was used to measure the segmentation errors by three metrics: segmentation error rate (ER), local consistency error (LCE), and bidirectional consistency error (BCE). The performance of a lossy data compression was introduced as an image segmentation method in color–texture natural images and to assess the performance measurements of the method, four metrics were used, that is, PRI, GCE, VOI, and BDE.[7] The graph-based image segmentation processes have a big problem: finding the reasonable weights for the graph cuts implementation, like the previous one, the PRI, GCE, VOI, and BDE metrics were used to compare an unsupervised multilevel color image segmentation method with others.[47] A hybrid roof segmentation approach was presented for the color aerial images in the work of El Merabet et al.[48] Its qualitative image segmentation was evaluated by the mean value of VINET from the work of Vinet.[49] Following that different segmentation evaluation methods are employed when dealing with different studies, that is, object accuracy measurement: Jaccard index (JI),[36,50–52] similarity matching methods: Dice coefficient (Dice),[36,37,51] image contour matching methods: Hausdorff distance (HAUSD),[22,50] and/or contour-based evaluation: normalized sum of distances (NSD);[53,54] pixel difference between a segmented image and the ground-truth (GT) image: Hamming distance (HD),[36,55,56] Fowlkes–Mallows index (FMI)[57–59] and boundaries discrepancy between them: boundary Hamming distance (BHD),[60] polyline distance metric (PDM);[61–63] surface distance measurement methods: mean absolute surface distance (MASD);[64–66] statistic based methods: mutual information (MI)[34,67] and/or normalized mutual information (NMI);[68,69] texture-based methods: earth movers distance (EMD);[70–72] relevance-based metrics: recall and precision.[73,74] The conceptual framework of image understanding and computer vision systems will be discussed in the following subsections.

### Conceptual Image and Vision Processing System Design

The overall framework of image analysis systems is shown in Fig. 1. Normally, the image acquisition stage is to enter the number of sequence images from a benchmark standard dataset or a real-time image capturing device. The image preprocessing stage can also be a preliminary stage of the systems, in which a raw image is normalized into desired image.[75] In the next stage, the desired image segmentation stage is to separate uniform regions in a desired image, and postprocessing may be used to enhance the regions. A set of local and global features from these regions of the desired image is extracted to describe the entrance image, which are used in the classification stage.

### Hierarchical Structure in Color Image Segmentation Evaluation

The extended hierarchy of the color image segmentation evaluation process in the work of Zhang et al.[42] is shown in Fig. 2; this process can be divided into three main categories: subjective, objective, and hybrid subjective–objective evaluations. A point of note is that the performance measurement of image segmentation methods must meet strict criteria and are compared with other studies. The details of main categories and/or subcategories are presented as follows:

- ***Subjective evaluation*:** This evaluation of image segmentation algorithms is a frequently used process for the measurement of image (video) quality. In general, there are two main determinant factors that contribute to a more efficient evaluation of the segmentation algorithms: different experimental conditions, and different human observation experiences and skills. However, subjective evaluation is a time-consuming, energy-intensive, and error-prone process, which can cause an exceedingly tedious task for a viewer. Furthermore and also noteworthy, a quantitative–subjective evaluation method cannot reproduce the statistical properties of the segmented images.[34,37,50]
- ***Objective evaluation*:** This evaluation of image segmentation is based on quantitative criteria and analyzed by internal observation. In fact, the objective methods calculate appropriate metrics directly from the images; these metrics try to measure certain similarities and significant differences between a segmented image and its GT image. To prove the quality validation of a segmentation method, it is necessary to avoid the influence of human activities and factors, to provide consistency, and also to improve reliability in segmenting different types of images.[39] However, this kind of segmentation evaluation has a number of limitations when utilizing the hand-segmenting benchmark datasets as follows: a variety of file formats for hand-segmenting images, different GT segmentation images for a given image, and the datasets are too large, thus providing hardware with necessary processing capabilities.
  - *System-level evaluation:* Different evaluation metrics attempt to assess the impact of a designed segmentation method on the complete system. A well-designed segmentation method provides useful information on tracking performance analysis of the system; hence, this task is needed to investigate the significant accuracy improvement on an applicative system. However, it is very difficult in many cases to use a large number of manually segmented images to accomplish fairly accurate pixel-level evaluation criteria. Therefore, obtaining high system accuracy shows that the proposed segmentation method also has a high-performance design. Three criteria in the work of Tomalik[76] are presented for the impact of a segmentation method on the performance analysis of the system classification: (1) minimizing and avoiding false positives, (2) fitting as much as possible the segmented regions into their expected hand-segmented regions, and (3) the segmentation findings would be the most useful for expansion and further analysis.
  - *Analytical evaluation:* These methods are considered as result-independent methods to be analyzed based on the principles, properties, for example, processing scheme and algorithmic structure, and the processing complexity of image segmentation methods.[34] In theory, the image segmentation methods can be assessed analytically.[77]
  - *Visual evaluation:* In these methods, a benchmark dataset is separated into different image groups with the same subject matter,[78,79] in which an evaluation method is statistically established by analyzing the individual and group-level differences between the segmented images and their original images. The main goal of these methods is to show the quantification of the quality of an image segmentation results that can display more detailed visual information. Another goal is to analyze possible errors in the segmentation stage. On the other hand, the difficult problem is that the diversity in the image region types of different image databases can lead to lower segmentation performance. However, the crucial challenge is that the visual evaluation is still under development.
  - *Supervised evaluation:* In order to evaluate the accuracy of supervised segmentation methods, the segmentation results are used to compare between GT images and segmented images. In fact, the methods are directly based on region overlapping information; in which reviewers tend to consider the possibility of the standard quality measurement methods.[51] The supervised segmentation evaluation can be also divided into two subgroups: area-based and edge-based evaluations.[80] Briefly, area-based evaluation methods are derived from the intersection of segmented image pixels and hand-segmenting image pixels. On the other hand, edge-based evaluation methods often provide greater insight into analyzing the image segmentation error between the image edges. A more significant discussion can be found in Refs..[51,54,80,81]
  - *Unsupervised evaluation:* The unsupervised evaluation of color image segmentation methods, objectively in a quantitative manner, can be designed for measuring the region-by-region correlation and differences of segmented images in observed original images; which means that the error between these images can be obtained without the reference-segmented images;[82] hence, it becomes harder due to little information concerning the best consensus GT for original image dataset. The main issues addressed in an efficient segmentation for different images are as follows: (1) textures and/or regions are homogeneous, in which neighboring regions must be readily distinguishable from each other; (2) to have simple region interiors; and (c) region boundaries must not lose real boundaries.[42,83,84]

- *Hybrid objective–subjective evaluation*: A hybrid objective–subjective evaluation process is employed to measure the performance of an image segmentation approach with perceptual analysis. The main goal of objective evaluation metrics is to assess the performance of the quality of images. Subjective evaluation is needed to measure the image quality and provide the human visual comparison of an image details. It can be useful in understanding how humans perceive visual quality of images which can influence the intricacy of visual information.

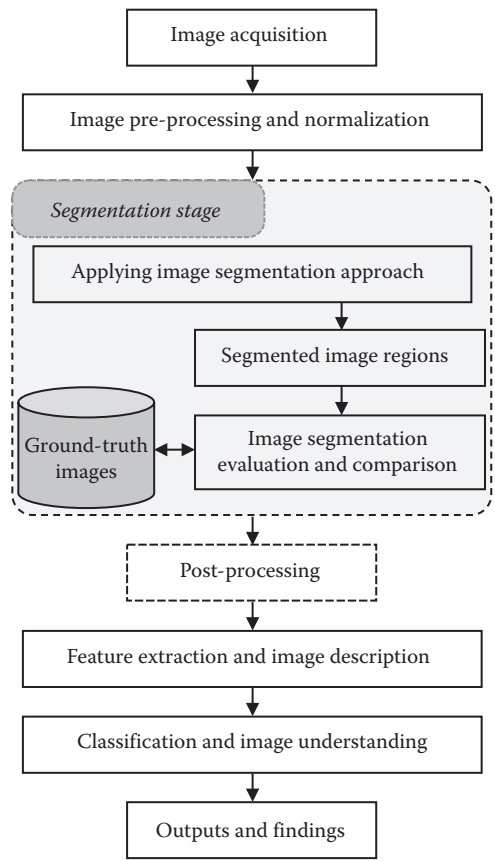

**Fig. 1** An overal framework of computer vision and image undestanding system

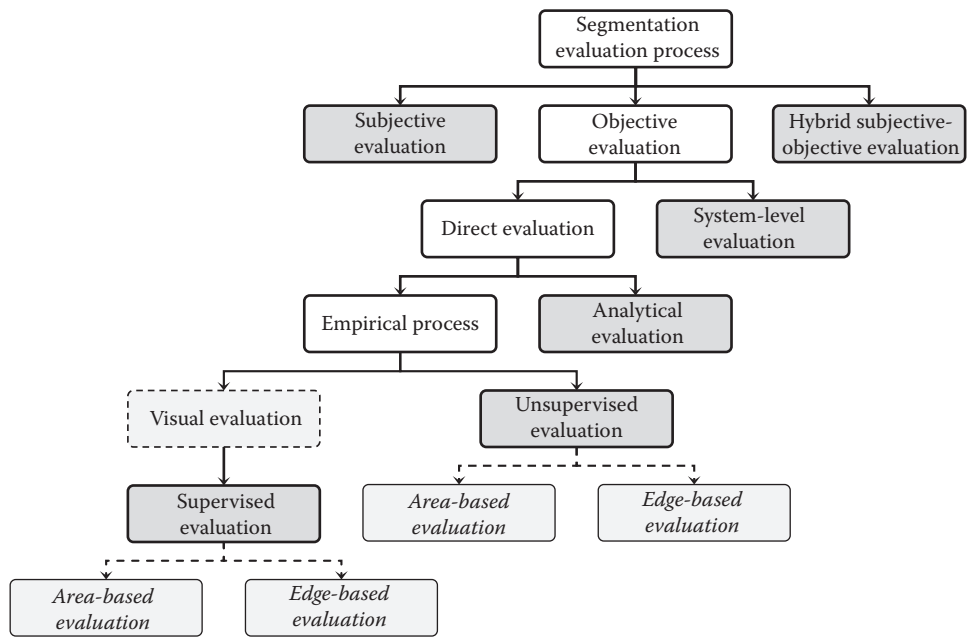

**Fig. 2** A hierarchical structure for color image segmentation evaluation

## TYPES OF QUANTITATIVE METRICS

Quantitative metrics are proposed to evaluate the performance of image segmentation methods using the intuitive knowledge. Typically, comparable segmentation results and/or errors will be necessary to help advance the state of the art. Therefore, the categorization of quantitative metrics into different performance groups is critical to support further development. The following subsections describe this categorization in more detail.

## Region/Volume Based

The region/volume based measures are based on the segmented object regions that can provide an evaluation of outline matching in pixel-to-pixel autosegmentation and its reference images. Rather than considering performance metrics, these measures could be divided into two subgroups: statistical/probabilistic analysis and distance analysis.

### Statistical/Probabilistic Analysis

In fact, the target of the statistical analysis is to provide a set of result values for comparison of a proposed segmentation method with other methods, where this work can calculate the overall probability of segmentation results for all image pixels.

a. Relevance measurement
The notion of relevance measurement is a computation sequence for analyzing, in which segmented image pixels have been detected correctly in their respective regions of origin, that is, tagged GT data, or have not been detected correctly. For image segmentation, a confusion matrix for this experiment is shown in Fig. 3. As shown in Fig. 4, the basic concepts of the confusion matrix are defined as follows:

— True positives denote the number of pixels, regions, and/or objects of the foreground of a segmented image, that is, region of interest, that are truly identified with its foreground GT image.

|  | | Automatic segmentation result | |
|---|---|---|---|
|  | | Positive | Negative |
| Ground truth | Positive | $TP = \sum_{i=1}^{k} n_{(i,i)}$ | $FN = \sum_{i=1}^{k}\sum_{j \neq i}^{k} n_{(i,j)}$ |
| | Negative | $FP = \sum_{j=1}^{k}\sum_{i \neq j}^{k} n_{(i,j)}$ | $TN = \sum_{j=1}^{k}\sum_{i \neq j}^{k}\sum_{j \neq i}^{k} n_{(i,j)}$ |

**Fig. 3**  A confusion matrix of segmented images and their GT images

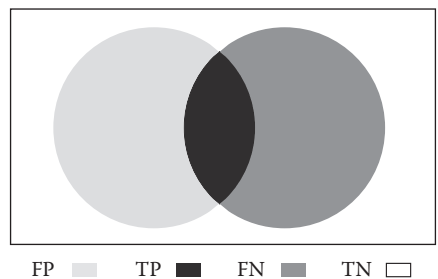

FP  TP  FN  TN

**Fig. 4**  The conception of confusion: segmented image pixels (left area), their respective GT image pixels (right area), and image background (white area)

- False positives denote the number of pixels, regions, and/or objects of the foreground of a segmented image that are falsely identified with its foreground GT image.
- True negatives denote the number of pixels, regions, and/or objects of the background of a segmented image that are truly identified with its background GT image.
- False negatives denote the number of pixels, regions, and/or objects of the background of a segmented image that are falsely identified with its background GT image.

Some well-known relevance metrics can be defined as presented in Table 1; these metrics are overlap based, so they indicate the ability of the system's performance to evaluate mis-segmented pixels and correct segmentation results, that is, target pixels.

The VD metric measured the volume of the foreground pixel regions on both segmented images and their reference images; to proof this point, after some algebraic manipulations, the Eq. VD is modified as follows:

$$\begin{aligned} VD &= \frac{|FN - FP|}{2TP + FN + FP} \\ &= \frac{\left\| \left( |S_{GT}(f)| - TP \right) - \left( |S_{auto}(f)| - TP \right) \right\|}{(FN + TP) + (FP + TP)} \\ &= \frac{\left\| |S_{GT}(f)| - \cancel{TP} - |S_{auto}(f)| + \cancel{TP} \right\|}{\underbrace{(FN + TP)}_{|S_{GT}(f)|} + \underbrace{(FP + TP)}_{|S_{auto}(f)|}} \quad (1) \\ &= \frac{\left\| |S_{GT}(f)| - |S_{auto}(f)| \right\|}{|S_{GT}(f)| + |S_{auto}(f)|} \quad \blacksquare \end{aligned}$$

b. Similarity measurement

*JJI*: In the case of segmentation evaluation, most researchers identify the significance of a comparative study of similarity measures. One of these metrics that computes the similarity between the GT images and

**Table 1** Quantitative relevance metrics derived from confusion matrix

| Evaluation metrics | Symbols/names | Equations |
|---|---|---|
| True Negative Rate | TNR, specificity | $TNR = \frac{TN}{TN + FP}$ (1) |
| True Positive Rate | TPR, recall, sensitivity | $TPR = \frac{TP}{TP + FN}$ (2) |
| Positive Likelihood Ratio | PLR | $PLR = \frac{Sensitivity}{1 - Specificity}$ (3) |
| Negative Likelihood Ratio | NLR | $NLR = \frac{1 - Sensitivity}{Specificity}$ (4) |
| False Positive Rate | FPR | $FPR = \frac{FP}{FP + TN}$ (5) |
| False Negative Rate | FNR | $FNR = \frac{FN}{FN + TP}$ (6) |
| Precision | $P$, Reliability | $P = \frac{TP}{TP + FP}$ (7) |
| F-measure | F | $F = \frac{Precision \times Recall}{Precision + Recall}$ (8) |
| XOR | XOR | $XOR = \frac{FP + FN}{TP + FN}$ (9) |
| Accuracy | AC, Validity | $AC = \frac{TP + TN}{TP + TN + FP + FN}$ (10) |
| Error Probability | EP | $EP = 1 - AC$ (11) |
| Volumetric Distance | VD | $VD = \frac{|FN - FP|}{2TP + FN + FP}$ (12) |
| Volumetric Similarity | VS | $VS = 1 - VD$ (13) |
| Area Under Curve | AUC | $AUC = 1 - \frac{FPR + FNR}{2}$ (14) |

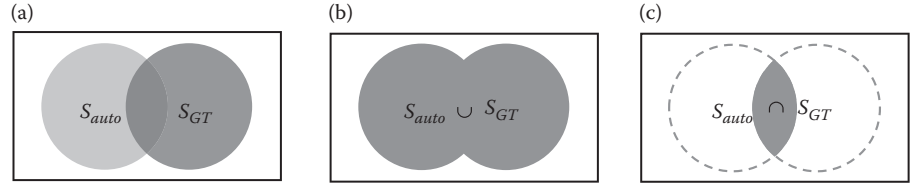

**Fig. 5** Symmetric differences: (a) overlapped areas; (b) union operation; (c) intersect operation

its automatic segmented result is the JI. This metric is defined as follows:

$$JI = \frac{|S_{auto} \cap S_{GT}|}{|S_{auto} \cup S_{GT}|} \quad (2)$$

where $S_{auto}$ is the automatic segmented images set, and $S_{GT}$ is its GT images set. The vertical line brackets denote the absolute value of set's elements; "$\cap$" and "$\cup$" symbols are the intersection and union operations, respectively (Fig. 5). With increase in the similarity between two sets, the number of elements in the intersection set will increase and the cardinality of union set will decrease. However, in considering this equation, when similarity of two sets reach maximum, the JI value becomes 1. And in similarity absence, JI will be 0. Often, this metric is applied when evaluating an automatic segmentation method on image dataset that consists of just two main regions: foreground and background.

*Dice coefficient*: The dice coefficient is a statistical-based metric that evaluates the performance of a color image segmentation method by comparing its result set with the hand-segmented image set. The mathematical calculation of Dice coefficient is presented as follows:

$$DICE = \frac{2 * |S_{auto} \cap S_{GT}|}{|S_{auto}| + |S_{GT}|} \quad (3)$$

The cardinality value of the sum of $S_{auto}$ and $S_{GT}$ sets is static, which means by increasing or decreasing the similarity of two sets, the denominator remains fixed; therefore, the value of the fraction increase with increase in the size of intersection set. However, the range of Dice coefficient value is between 0 and 1, where the values denote the relative effect of dissimilarity and similarity pixels between two sets, respectively. Following this, the XOR metric is employed in calculating the mis-segmented pixels probability,[85,86] which can be differently defined as follows:

$$XOR = \frac{|S_{auto} \cup S_{GT}| - |S_{auto} \cap S_{GT}|}{|S_{GT}|} s \quad (4)$$

*FMI*: It has been presented essentially for computing the similarity between two clustering results; but in image segmentation evaluation, it intends to compare two segmented areas by Eq. 5. The value range of this metric is in interval [0 1], where 0 and 1 denote the mismatching and the perfect agreement between two segmented images, respectively.

$$FMI = \frac{|S_{auto} \cap S_{GT}|}{\sqrt{|S_{auto}| \cdot |S_{GT}|}} \quad (5)$$

*Rand index* (RI): It was originally presented for clustering quality evaluation. This measurement works based on the point pairs in cluster placement. In the segmentation evaluation process, both the point pairs and the clusters are replaced with the pixel pairs and the segments, respectively. If any pixel pair in automatic segmentation results is placed in the same segment, the corresponding pixels in its human-segmented image must be in the same region as well; also, if a specific pixel pair in one belongs to separated segments, the corresponding pixels must be in separated segments; in this term, the similarity will be increased. However, the RI is defined in Eq. 6. The RI value range is between 0 and 1.

$$RI(S_{auto}, S_{GT}) = \frac{1}{\binom{N}{2}} \sum_{\substack{i,j=1 \\ i \neq j}}^{N} \mathbb{IF}\left[\left(P_i^{auto}, P_j^{auto}\right), \left(P_i^{GT}, P_j^{GT}\right)\right] \quad (6)$$

where $P_i^{auto}, P_j^{auto}$ and $P_i^{GT}, P_j^{GT}$ are pixel pairs in automatic segmented images and the corresponding pair in the GT images; also, $\mathbb{IF}$ is identity function that is defined as follows:

$$\mathbb{IF}\left[\left(P_i^{auto}, P_j^{auto}\right), \left(P_i^{GT}, P_j^{GT}\right)\right]$$

$$= \begin{cases} 1 & if \begin{cases} \left(\left(P_i^{auto} = P_j^{auto}\right) AND \left(P_i^{GT} = P_j^{GT}\right)\right) \\ OR \\ \left(\left(P_i^{auto} \neq P_j^{auto}\right) AND \left(P_i^{GT} \neq P_j^{GT}\right)\right) \end{cases} \\ 0 & Otherwise \end{cases}$$

$$(7)$$

*PRI*: Some image segmentation datasets are supervised, which have at least one GT segmented image for any existing image; in some cases, there are more than one GT for a given image. The RI is not appropriate for comparison of segmentation result with multiple human segmented images. Considering this condition, a modified RI called PRI is defined as follows:

$$PRI = \frac{1}{K}\sum_{j=1}^{K} RI\left(S_{auto}, S_{GT}^{j}\right) \quad (8)$$

where $K$ is the number of existing GT segmented images of a given image, in which $S_{GT}^{j}$ is the $j$th GT of this image. The range values of the segmentation results and their meaning are similar to RI metric.

*NPR*: Considering that the different aspects of image complexity is significant, it addresses its difficulty in evaluating segmentation result and describing more detailed information. In these cases, it is required to perform a more accurate evaluation of segmentation results by comparing them with different evaluation methods in a dataset. One of the ways to achieve this goal is to normalize the evaluation results by computing a baseline model; this model expresses the expected performance evaluation of an image segmentation method for a given dataset. A normalized evaluation that is presented in the work of Unnikrishnan et al.[87] is shown as follows:

$$NPR = \frac{PRI - EV}{MaxPR - EV} \quad (9)$$

where *PRI* is the current automatic segmented image, *MaxPR* is the maximum achieved *PRI* by applying the segmentation method through current image dataset, and the EV is computed as follows:

$$EV = \frac{1}{m}\sum_{j=1}^{m} PRI\left(S_{j}\right) \quad (10)$$

where $m$ is the number of images in a benchmark dataset, and $S$ denotes the automatic segmentation result of an image in the dataset.

*Misclassification errors*: This metric is introduced by Yasnoff et al.[88] to compute the percentage of displaced pixels in a given automatic segmented image by Eq. 11, regarding its GT image.

$$MCE = 1 - \frac{\left|S_{auto}(b) \cap S_{GT}(b)\right| + \left|S_{GT}(f) \cap S_{auto}(f)\right|}{\left|S_{auto}(b)\right| + \left|S_{auto}(f)\right|} \quad (11)$$

where $S_{auto}(b)\left(S_{auto}(f)\right)$ and $S_{GT}(b)\left(S_{GT}(f)\right)$ denote the background (foreground) pixels in the segmented and the GT images. The MCE value is in interval [0, 1], and it is similar to the other error metrics, lower value will be desirable.

*Error rate*: The error simply expresses the percent of pixels placed in a wrong region as shown in the following equation:

$$ER = \frac{N_{false} + N_{miss}}{N_{total}} \times 100 \quad (12)$$

where $N_{false}$ and $N_{miss}$ are the numbers of segmented pixels that have been located in the erroneous region, and $N_{total}$ denotes the number of pixels in the hand-segmented image. In these terms, the value range of ER is in the interval between 0 and 100, where 0 value shows there is no similarity between segmentation results and corresponding GT image; also the value of 100 denotes the excellent segmentation results.

*MI*: It is a metric that analyzes the relationships between two sets of variable regions by measuring the amount of feature information of a set associated with the other set as expressed in Eq. 13 and presented in Refs.[89,90]

$$MI(S_{auto}, S_{GT}) = Ent(S_{auto}) + Ent(S_{GT}) - Ent(S_{auto}, S_{GT}) \quad (13)$$

where $Ent\left(S_{auto}\right)$ and $Ent\left(S_{GT}\right)$ denote the marginal entropy, and $Ent\left(S_{auto}, S_{GT}\right)$ is the joint entropy.

*VOI*: This metric has been basically proposed for clustering comparison. It can denote the measurement of missed and extra information rates in resemblance of the two image sets, which is based on their mutual information and marginal entropy as follows:

$$VOI(S_{auto}, S_{GT}) = Ent(S_{auto}) + Ent(S_{GT}) - 2MI(S_{auto}, S_{GT}) \quad (14)$$

*NMI*: As described earlier, the MI measures the amount of shared information between two image sets; but because of the upper bound problem, it seems to be necessary to normalize this metric to achieve a fixed maximum value as an upper bound limit. Eq. 15 shows how Strehl and Ghosh[91] introduced the NMI for solving this problem:

$$NMI = \frac{MI\left(S_{auto}, S_{GT}\right)}{\sqrt{Ent\left(S_{auto}\right).Ent\left(S_{GT}\right)}} \quad (15)$$

c.  Consistency error measurements

Some quantitative evaluation methods are used for measuring the consistency between automatic segmented image and its GT image. It is suggested that there are three measures, that is, GCE, LCE, and BCE; to calculate them, it is necessary to compute local refinement error (LRE) as nonsymmetric formula in the following equation:

$$LRE(S_{auto}, S_{GT}, P) = \frac{\left|R(S_{auto}, P) \setminus R(S_{GT}, P)\right|}{\left|R(S_{auto}, P)\right|} \quad (16)$$

where $R(S_{auto},P)$ denotes a region in segmented image that contain pixel $P$, and "\" symbolizes set difference. The meaning of other variables and operators are the same as previously mentioned. Now, LCE, GCE, and BCE can be computed. The value range of these metric is between 0 and 1, in which 0 value represents no mis-segmented region and 1 value means that there is no homogeny region between two image sets.

*LCE*: Some metrics provide the label refinement homogeneously through the image in order to rectify slight difference in graininess when comparing segmentations. The LCE is a measurement (Eq. 17) that shows the degree of refinement between automatic segmentation result and its GT.

$$LCE = \frac{1}{N}\sum_{i=1}^{N} MIN\{LRE(S_{auto},S_{GT},P_i), LRE(S_{GT},S_{auto},P_i)\}$$

(17)

where $N$ is the total number of image pixels.

*GCE*: In this measurement, it can be assumed that any segmentation like $S_i$ possesses a refinement of segmentation $S_j$ (Eq. 18). As shown in Eq. 14, this metric reports maximum similarity, when all local refinements are placed in the same direction. It is shown that in the absence of accommodation, no similarity will be achieved.

$$GCE = \frac{1}{N} MIN\{\sum_{i=1}^{N} LRE(S_{auto},S_{GT},P_i), \sum_{i=1}^{N} LRE(S_{GT},S_{auto},P_i)\}$$

(18)

*BCE*: When excellent segmentation refinement is achieved, it is necessary to measure the degree of refinement in order to provide more accurate quantitative analysis; in this situation, one may simply set the minimal LRE to a very low value, so that by enhancing segmentation method, the LCE will reduce slowly. The BCE is applied when LCE is very low, it is defined as follows:

$$BCE = \frac{1}{N}\sum_{i=1}^{N} MAX\{LRE(S_{auto},S_{GT},P_i), LRE(S_{GT},S_{auto},P_i)\}$$

(19)

## Distance-Based Metrics

Accurate region contour is a key point in image segmentation evaluation; hence, some image segmentation measurements focus on distance between segmented object's contour and reference object's boundary. Some metrics work on physical distance[92,93] and few are based on logical distance.[94,95] In physical distance, there are several geometric distance measurement methods, for example, Euclidean distance, Manhattan distance, etc. And in logical distance, the value of distance can only be 0 or 1, where 0 means there is no distance between two objects and 1 shows that there is a distance. However, the most used physical distance measurement method is two-dimensional Euclidean distance, which is defined as follows:

$$d_{EQ}(P_1,P_2) = \sqrt{(x_{p1}-x_{p2})^2 + (y_{p1}-y_{p2})^2}$$

(20)

where $P_1(x_{p1},y_{p1})$ and $P_2(x_{p2},y_{p2})$ are the first and second points, respectively; also, a derived concept from Euclidean distance is a distance distribution signature (the so-called minimal Euclidean distance as shown in Eq. 21) that shows minimum Euclidean distance between a point and a set of points.

$$d_{min}(p_s,B_t) = MIN\{d_{EQ}(p_s,p_t)|p_t \in B_t\}$$

(21)

where $p_s$ and $p_t$ symbolize the pixels that are located on the boundary of automatic segmented object and its GT boundary, respectively. Another derived concept is supermum–infimum distance that shows minimum Euclidean distance between two furthest points from two distinct images sets:

$$d_{sup-inf}(B_{auto},B_{GT}) = MAX\{d_{min}(p_i,B_{GT})|p_i \in B_{auto}\}$$

(22)

where $B_{auto}$, and $B_{GT}$ denote the automatic segmented object boundary and its human segmented GT boundary sets, respectively. It is citable that minimal Euclidean and supermum–infimum distance are not symmetric.

*HAUSD*: This metric measures the longest Euclidean distance between two sets of boundary pixels from automatic segmented and human segmented objects; its definition is given by the following equation:

$$HAUSD = MAX\{d_{sup-inf}(B_{auto},B_{GT}), d_{sup-inf}(B_{GT},B_{auto})\}$$ (23)

It is obvious that 0 value for the HAUSD shows that there is exhaustive matching between result and its reference. With increasing mismatching between two sets, the distance will increase too.

*MASD*: As explained earlier, the minimal Euclidean distance is not symmetric, because the minimal Euclidean distance from $B_{GT}$ to $B_{auto}$ is not equal to the distance from $B_{auto}$ to $B_{GT}$; hence, to show the average of distances between two boundary pixel sets, the mean absolute surface distance was introduced in the following equation:

$$MASD = \frac{1}{2}\left(\frac{1}{N}\sum_{p_i \in B_{auto}}^{N} d_{min}(p_i,B_{GT}) + \frac{1}{M}\sum_{p_j \in B_{GT}}^{M} d_{min}(p_j,B_{auto})\right)$$

(24)

In considering this equation, the minimum value for the metric is 0 which means superior matching is achieved but because of losing upper bound for mean absolute surface distance, there is no maximum value to denote the explicit dissimilarity.

*NSD*: The metric provided the basis for computing the distance between any misclassified pixel in automatic segmentation result and the contour of its referenced object's GT. To normalize the measurement, the sum of these distances is divided by the sum of the distances between any pixel in the result and the boundary of its reference. The NSD equation is defined as follows:

$$NSD = \frac{\sum_{p_i \in \{S_{auto} \triangle S_{GT}\}} d_{\min}(p_i, B_{GT})}{\sum_{p_j \in \{S_{auto} \cup S_{GT}\}} d_{\min}(p_j, B_{GT})} \quad (25)$$

Like other distance-based metrics, a lower value for distance denotes the perfect segmentation and a higher value refers to the undesired segmentation results. Hence, after the metric is normalized, the value range will be between 0 and 1.

*Average Symmetric Surface Distance* (*ASD*): Another distance-based metric is based on the ASD relating to two image segmentation results, which reports the average of the minimum Euclidean distance between the contours of the corresponding objects in both segmentations as follows:

$$ASD = \frac{\sum_{p_j \in B_{GT}} d_{\min}(p_j, B_{auto}) + \sum_{p_i \in B_{auto}} d_{\min}(p_i, B_{GT})}{|B_{auto}| + |B_{GT}|} \quad (26)$$

Obviously, the lesser value of ASD denotes more desirable segmentation results.

*BDE*: One of the most distinctive quantitative segmentation evaluation methods is the BDE that is focused on the distance-based features, where this metric is sensitive to the distance between the contours of an object in the automatic segmented image and the contours of the corresponding object in its GT image. In other words, it computes the average of the distance between any pixel in the boundary of an automatic segmented object and the nearest pixel in the contour of the reference object as follows:

$$BDE = \frac{1}{N} \sum_{p_i \in B_{auto}} d_{\min}(p_i, B_{GT}) \quad (27)$$

where *N* is the number of the boundary pixels in an automatic segmented image. The lower bound of BDE is 0 which denotes perfect matching between the contours of both automatic segmented and GT; although, there is no upper bound, any high value means undesirable segmentation result.

*HD*: It is a metric based on logical distance originally presented to calculate the disagreements between two strings. In segmentation evaluation, it is used to enumerate the number of displaced pixels as shown in Eq. 28. To normalize the distance, it is divided by the total number of pixels; in this condition, the value range is given as [0, 1], where 0 and 1 denote absolute dissimilarity and excellent matching between a segmentation method result on an image and its GT.

$$HD = \frac{|S_{auto}(b) \cap S_{GT}(f)| + |S_{GT}(b) \cap S_{auto}(f)|}{|S_{auto}|} \quad (28)$$

*BHD*: This metric is a variation of HD that processes to the determination of the contour of automatic segmented object. When an accurate object boundary is involved, the common HD is not applicable anymore. The BHD calculates the mis-segmented boundary pixel on original boundary pixels in Eq. 29. The value range and its description are similar to the HD.

$$BHD = \frac{|B_{auto}(b) \cap B_{GT}(f)| + |B_{GT}(b) \cap B_{auto}(f)|}{|B_{GT}(f) \cup B_{auto}(f)|} \quad (29)$$

## DISCUSSION AND SUGGESTION

The system-level evaluation of a proposed image segmentation method can show the impact of the designed method on a complete image analysis system; also, the outcome of the segmentation evaluation results will set a high standard for automatic qualitative evaluation, and it may provide desirable features of a precedent for future enhancements. In many situations, it is difficult to choose a set of segmentation metrics for a fair comparison with other existing methods. In order to show a state-of-the-art segmentation accuracy of the standard benchmark test datasets, it is necessary to consider and investigate the morphological features of the regions of the test images, where these regions are formed in different textural structures, natural conditions, and colors. The most commonly used metrics are summarized in Table 2.

There has been much effort to provide a common potential explanation and to find the best self-balancing stability among the selected metrics for the accuracy of a proposed image segmentation method, for example, natural image segmentation methods with respect to. PRI and VOI metrics;[96] unsupervised image segmentation methods with respect to PRI, VOI, GCE, and BDE;[47,97] parallelized segmentation methods with respect to HM and NSD; automatic brain tumor segmentation methods with respect to Dice and EP; and nuclei segmentation methods with respect to RI, JI, HAUSD, and NSD.[54] As shown in Table 3, the desired values for all the mentioned metrics are summarized, which means the direction of mean changes for each metric: higher means better ("↑") and smaller means better ("↓").

**Table 2** Image segmentation evaluation methods

| Metric name | Symbol | Region based | | | Boundary based | | | Binary segmentation | General segmentation | Brief description |
|---|---|---|---|---|---|---|---|---|---|---|
| | | Distance | Relevance | Similarity | Distance | Relevance | Similarity | | | |
| True Negative Rate (Specificity) | TNR | | ✓ | | | | | ✓ | | A set of pixels/voxels as a nonobject |
| True Positive Rate (Recall, Sensitivity) | TPR | | ✓ | | | | | ✓ | | A number of correct overlapping pixels/voxels |
| Positive Likelihood Ratio | PLR | | ✓ | | | | | ✓ | | A subsegment location probability rate |
| Negative Likelihood Ratio | NLR | | ✓ | | | | | ✓ | | A non-subsegment location probability rate |
| False Positive Rate | FPR | | ✓ | | | | | ✓ | | A set of autosegment pixels/voxels labeled as nonobject |
| False Negative Rate | FNR | | ✓ | | | | | ✓ | | A set of hand-segment pixels/voxels labeled as nonobject |
| Precision | P | | ✓ | | | | | ✓ | | A measure of segmentation accuracy |
| Accuracy | AC | | ✓ | | | | | ✓ | | A ratio of correct overlapping over the number of all autosegment pixels/voxels |
| XOR | XOR | | ✓ | | | | | ✓ | | A ratio of the pixels/voxels mis-segmented over the number of all hand-segment pixels/voxels |
| F-measure | F | | ✓ | | | | | ✓ | | A harmonic mean of precision and recall rates |
| Error Probability | EP | | ✓ | | | | | ✓ | | The probability of mis-segmenting an object |
| Volumetric Distance | VD | | ✓ | | | | | ✓ | | The mean absolute volume difference between hand-autosegment and the sum of their volumes |
| Volumetric Similarity | VS | | ✓ | | | | | ✓ | | A ratio of volumes of the hand-autosegments to show the similarity of them |
| Area Under Curve | AUC | | ✓ | | | | | ✓ | | A ratio of the segmentation performance that how well the correct and mis-segment are |
| Jaccard Index | JI | | | ✓ | | | | ✓ | | A ratio of how closely the autosegment overlaps the hand-segment pixels/voxels |
| Dice Coefficient | Dice | | | ✓ | | | | ✓ | | A ratio of overlapping over the average size of both hand and autosegment pixels/voxels |
| Fowlkes-Mallows Index | FMI | | | ✓ | | | | ✓ | | A geometric mean of precision and recall rates |

(*Continued*)

Table 2  Image segmentation evaluation methods (*Continued*)

| Metric name | Symbol | Region based | | | Boundary based | | | Binary segmentation | General segmentation | Brief description |
|---|---|---|---|---|---|---|---|---|---|---|
| | | Distance | Relevance | Similarity | Distance | Relevance | Similarity | | | |
| Rand Index | RI | | | ✓ | | | | ✓ | ✓ | A measure of the similarity between hand and autosegment pixels/voxels |
| Probabilistic Rand Index | PRI | | | ✓ | | | | ✓ | ✓ | A measure of the similarity between auto-segment and a set of hand-segment pixels/voxels |
| Normalized Probabilistic Rand Index | NPR | | | ✓ | | | | ✓ | ✓ | Measuring of similarity between autosegment methods and a set of their hand segments |
| Misclassification Errors | MCE | | ✓ | | | | | ✓ | | Known as the similar mis-segmentation errors |
| Error Rate | ER | | ✓ | | | | | ✓ | ✓ | The ratio of the counts of the incorrect auto-segment pixels/voxels to the reference segments |
| Local Consistency Error | LCE | | | ✓ | | | | ✓ | ✓ | A ratio of identical segmentations |
| Global Consistency Error | GCE | | | ✓ | | | | ✓ | ✓ | A measure for ignoring over-segmentations |
| Bidirectional Consistency Error | BCE | | | ✓ | | | | ✓ | ✓ | An approximation ratio for a minimization of a fit error over a set of hand segments |
| Mutual Information | MI | | | ✓ | | | | ✓ | ✓ | A measure based on the concept of conditional entropy |
| Variation of Information | VOI | | | ✓ | | | | ✓ | ✓ | A measure based on the concept of both conditional entropy and mutual information |
| Normalized Mutual Information | NMI | | | ✓ | | | | ✓ | ✓ | Known as the dissimilarity or distance measure |
| Hausdorff Distance | HAUSD | | | | ✓ | | | ✓ | ✓ | A measure of proximity between the auto-segment pixels/voxels and its hand segment |
| Mean Absolute Surface Distance | MASD | | | | ✓ | | | ✓ | ✓ | A measure for how much on average the two autosegment and its hand segment differ |
| Average Symmetric Surface Distance | ASD | | | | ✓ | | | ✓ | ✓ | A measure for how close two segmented surfaces/contours are |
| Normalized Sum of Distances | NSD | ✓ | | | ✓ | | | ✓ | ✓ | A ratio of measuring the final distance between both hand and autosegment pixels/voxels |
| Boundary Displacement Error | BDE | | | | ✓ | | | ✓ | ✓ | A displacement error measurement between nearest hand and autosegment boundaries |
| Hamming Distance | HD | ✓ | | | | | | ✓ | | The number of pixels/voxels in disagreement of two hand and autosegment sets |
| Boundary Hamming Distance | BHD | | | | ✓ | | ✓ | ✓ | | A measure based on the concept of both Hamming distance and the object boundaries |

**Table 3** Balancing relationship values in the segmentation metrics

| TNR and TPR | | | | | | | | | |
|---|---|---|---|---|---|---|---|---|---|
| ↑↓ | | | | | | | | | FPR, FNR |
| ↑↑ | ↓↑ | | | | | | | | P, AC |
| ↑↓ | ↓↓ | ↑↓ | | | | | | | XOR |
| ↑↑ | ↓↑ | ↑↑ | ↓↑ | | | | | | F |
| ↑↓ | ↓↓ | ↑↓ | ↓↓ | ↑↓ | | | | | EP, VD |
| ↑↑ | ↓↑ | ↑↑ | ↓↑ | ↑↑ | ↓↑ | | | | AUC, VS, JI, Dice, FMI, RI, PRI, NPR |
| ↑↓ | ↓↓ | ↑↓ | ↓↓ | ↑↓ | ↓↓ | ↑↓ | | | MCE, ER, LCE, GCE, BCE |
| ↑↑ | ↓↑ | ↑↑ | ↓↑ | ↑↑ | ↓↑ | ↑↑ | ↓↑ | | MI |
| ↑↓ | ↓↓ | ↑↓ | ↓↓ | ↑↓ | ↓↓ | ↑↓ | ↓↓ | ↑↓ | VOI |
| ↑↑ | ↓↑ | ↑↑ | ↓↑ | ↑↑ | ↓↑ | ↑↑ | ↓↑ | ↑↑ | ↓↑ NMI |
| ↑↓ | ↓↓ | ↑↓ | ↓↓ | ↑↓ | ↓↓ | ↑↓ | ↓↓ | ↑↓ | ↓↓ ↑↓ HAUSD, MASD, ASD, NSD, BDE, HD, BHD |

Figure 6 illustrates a set of sample images, in which the objects were located and identified; there are possible changes in the autosegmented object location and structure that are intentionally placed within the location of the GT object. It should be noted that segmentation evaluation results are often based on the size and location of overlapping object areas, specifically in similarity-based computational methods; thus, different object areas and shapes with fixed-pixel sizes, that is, the autosegmented object (1,225 pixels), the hand-segmented object (4,900 pixels), and the entire background image (10,000 pixels), are drawn; and their average pixel intensity values are measured and compared in six different test sample images as shown in Table 4. Computation of the relevance metrics of such test images are presented in Fig. 7, where according to Table 3, the relationship values among the segmentation results are a quantitative representation of the expected trend. Due to the common boundaries in some sample images, that is, S3 and C3, the BHD is different from others (the gray cells in Table 4); the best evaluation results would be achieved when the number of common boundary pixels in both a machine-segmented object and its reference region are similar.

In order to investigate how the overlapped pixels, that is, intersection of the reference object and machine-segmented object, can influence the segmentation evaluation results, where the number of object pixels are approximately identical, a set of sample images associated with different rotation sequences are assumed and presented in Fig. 8. As shown in Fig. 9, the performance relevance measurements indicate that an object similarity interpretation of the segmentation method is based on the shapes and locations of both objects. Furthermore, Table 5 presents comparison of the segmentation evaluation results which shows the best scores depend on the equivalent fraction of the geometrical properties in the objects.

For another evaluation of color image segmentation methods, the original images with their four different hand-segmented images are shown in Fig. 10,[98] where there are value judgments in comparing these different images and the original image. As shown in Fig. 11, the proposed measures represent how much the human perceptions are fuzzy; to prove this, assume that three of the hand-segmented images are known as machine-segmented images (MSIs) with different number of segments. Both RI and PRI measures are approximately similar for both images, in which the highest value of these metrics is obtained when an MSI contains almost all the boundaries in the hand-segmented image and/or both the test and reference images have the same large regions. The smaller value of the VOI presents a segmentation closer to its reference image (Table 6). Moreover, The GCE metric is used for evaluating a measure of overlap between image pixels of the two segmentations, and it can be defined as the minimization of the image error between the two segmentations.

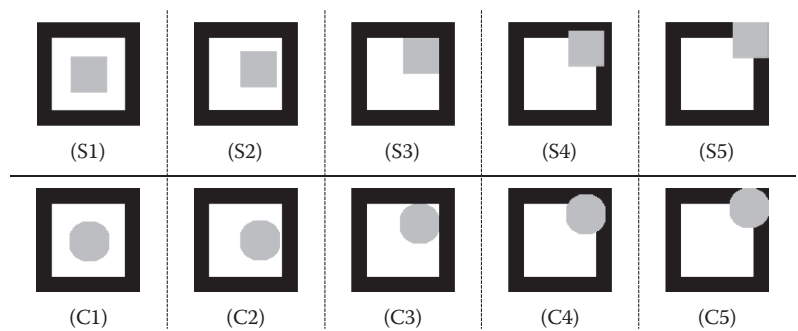

**Fig. 6** Test sample images set: an automatic segmented area (gray), its hand-segmented area (white), and its image background (black)

**Table 4** Performance evaluation results for the test sample images set

| Image name | JI | Dice | FMI | RI | VOI | GCE | BHD | TNR | TPR | FNR | P | F | XOR | AC |
|---|---|---|---|---|---|---|---|---|---|---|---|---|---|---|
| S1, C1, S2, C2 | 0.25 | 0.40 | 0.50 | 0.54 | 1.26 | 0.18 | 1.00 | 1.00 | 0.25 | 0.75 | 1.00 | 0.20 | 0.75 | 0.63 |
| S3 | 0.25 | 0.40 | 0.50 | 0.54 | 1.26 | 0.18 | 0.74 | 1.00 | 0.25 | 0.75 | 1.00 | 0.20 | 0.75 | 0.63 |
| C3 | 0.25 | 0.40 | 0.50 | 0.54 | 1.26 | 0.18 | 0.85 | 1.00 | 0.25 | 0.75 | 1.00 | 0.20 | 0.75 | 0.63 |

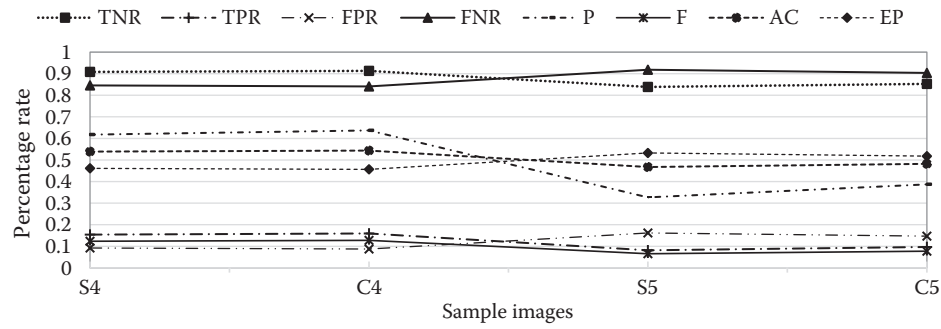

**Fig. 7** The relevance measurements of segmentation results

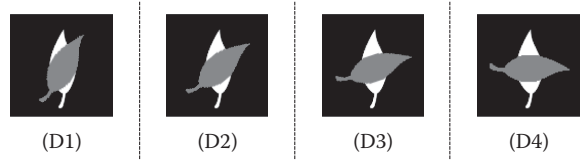

**Fig. 8** Test sample images: an automatic segmented area (gray), its hand-segmented area (white), and its image background (black) with the same volumetric areas

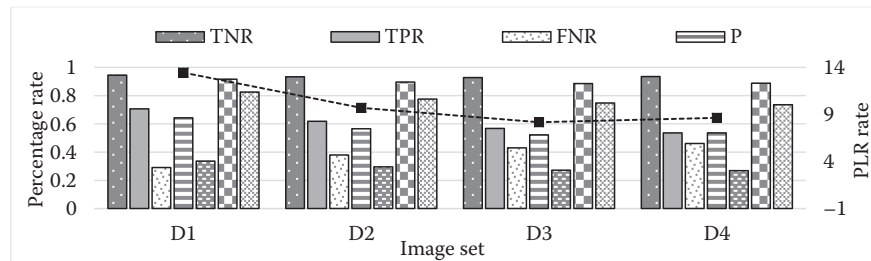

**Fig. 9** The segmentation results of various relevance measurements

**Table 5** Comparison of the evaluation results for the test sample images

| Image name | JI | Dice | FMI | RI | VOI | GCE | HAUSD | BHD |
|---|---|---|---|---|---|---|---|---|
| D1 | 0.51 | 0.67 | 0.68 | 0.85 | 0.70 | 0.13 | 19.10 | 0.34 |
| D2 | 0.42 | 0.59 | 0.59 | 0.82 | 0.80 | 0.15 | 23.71 | 0.42 |
| D3 | 0.37 | 0.55 | 0.55 | 0.80 | 0.85 | 0.16 | 26.08 | 0.45 |
| D4 | 0.37 | 0.54 | 0.54 | 0.81 | 0.83 | 0.16 | 26.00 | 0.46 |

## CONCLUSION

In this entry, a survey of performance evaluation metrics for color image segmentation methods is given, which can be used as a starting point for the selection of the most appropriate metrics for assessing automatic image segmentation procedure. In addition, a set of comparative results for different metric groups is presented. The key finding of this comparative study is that the balancing relationships among the metrics are mutually interacted. Besides, the metrics have been categorized from two viewpoints: the first is focused on the importance of volume and boundary sizes; the second is focused on the features of the segmentation results, for example, relevance and/or overlapping, and distance and similarity between autosegmented image results and its reference image. Moreover, a potential benefit of analyzing the segmentation evaluation results is the ability to characterize the system development process.

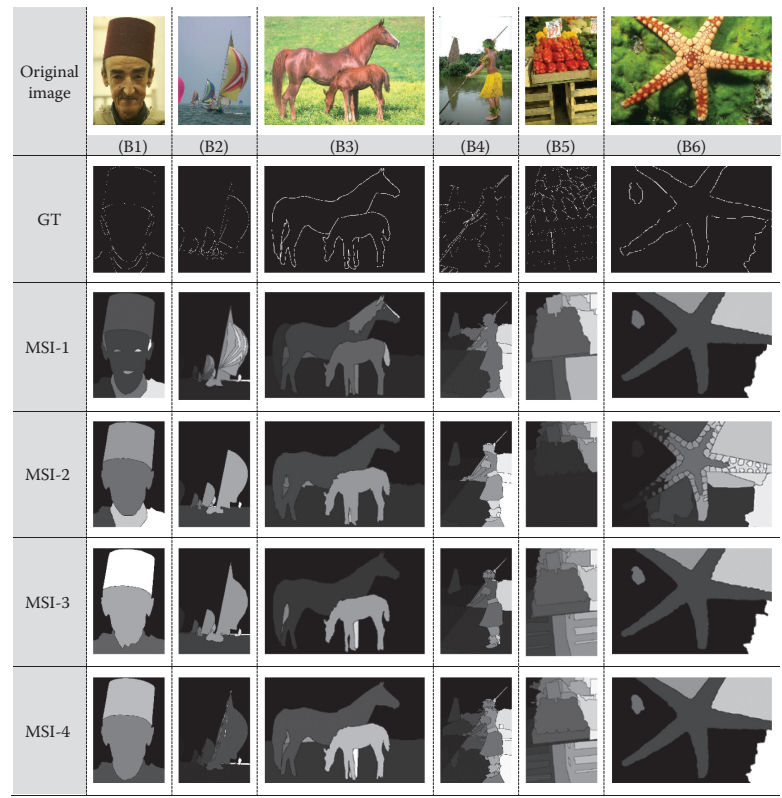

**Fig. 10** Examples of the GT and original images; (B1–B6) assumed machine-segmented methods (MSI)

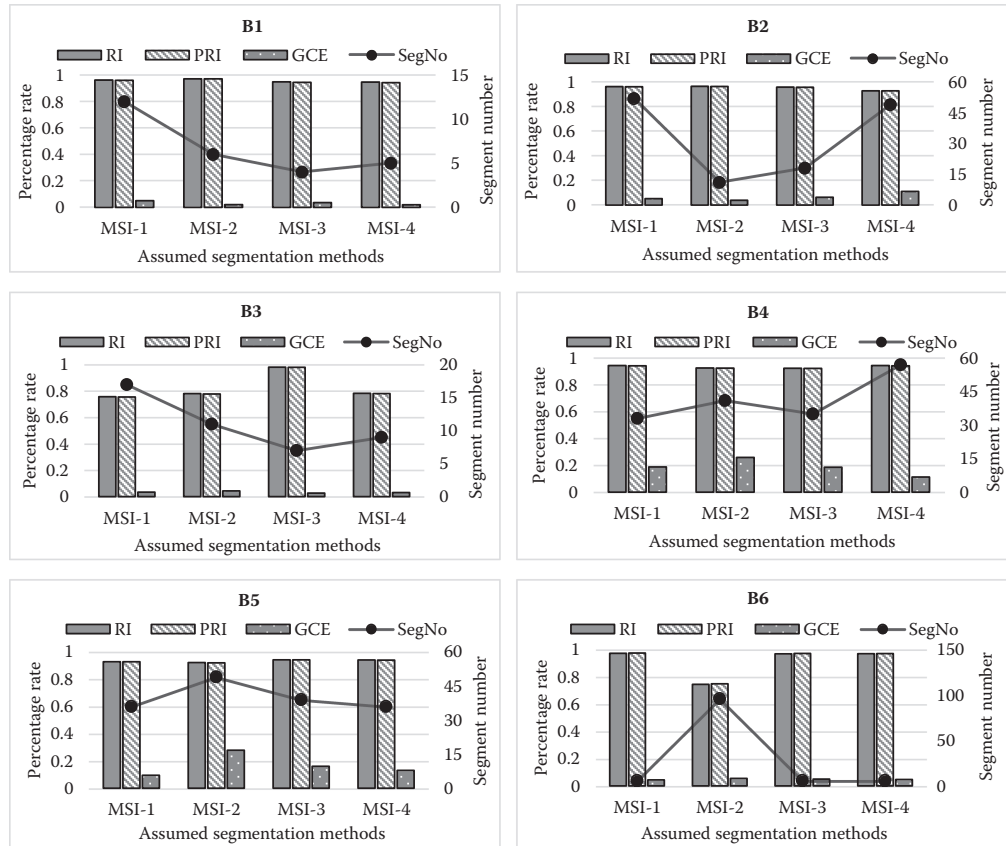

**Fig. 11** Examples of the original images; (B1–B6) and assumed machine-segmented methods (MSI)

Table 6  The evaluation results for the test sample images

| Methods | B1 | | | B2 | | | B3 | | | B4 | | | B5 | | | B6 | | |
|---|---|---|---|---|---|---|---|---|---|---|---|---|---|---|---|---|---|---|
| | MI | NMI | VOI | MI | NMI | VOI | MI | NMI | VOI | MI | NMI | VOI | MI | NMI | VOI | MI | NMI | VOI |
| MSI-1 | 1.69 | 0.91 | 0.35 | 3.39 | 0.77 | 2.05 | 2.58 | 0.81 | 1.25 | 1.25 | 0.66 | 1.28 | 1.72 | 0.81 | 0.83 | 2.12 | 0.89 | 0.55 |
| MSI-2 | 1.69 | 0.55 | 2.75 | 3.61 | 0.74 | 2.54 | 2.49 | 0.74 | 1.73 | 1.22 | 0.72 | 0.94 | 1.57 | 0.86 | 0.50 | 2.12 | 0.91 | 0.44 |
| MSI-3 | 1.67 | 0.90 | 0.38 | 3.74 | 0.80 | 1.90 | 2.39 | 0.75 | 1.57 | 1.25 | 0.91 | 0.24 | 1.54 | 0.83 | 0.64 | 1.83 | 0.83 | 0.75 |
| MSI-4 | 1.68 | 0.90 | 0.36 | 3.63 | 0.79 | 1.93 | 2.83 | 0.78 | 1.56 | 1.25 | 0.73 | 0.91 | 1.50 | 0.77 | 0.89 | 1.85 | 0.84 | 0.70 |